\pdfoutput=1

\documentclass[11pt]{article}

\usepackage[]{emnlp2022}

\usepackage{times}
\usepackage{latexsym}

\usepackage[T1]{fontenc}

\usepackage[utf8]{inputenc}
\usepackage{pifont}
\newcommand{\cmark}{\ding{51}}%

\usepackage{microtype}

\usepackage{amsmath,amsfonts,amssymb,bm}
\usepackage{pifont} 
\usepackage{graphicx,epsfig,fancybox} 
\usepackage{float}
\usepackage{color} 
\usepackage[ruled,vlined]{algorithm2e} %
\usepackage{algorithmic}
\usepackage{multirow}
\usepackage{comment}
\usepackage{hyperref}
\usepackage{natbib} 
\usepackage{subfigure}

\renewcommand{\vec}[1]{\boldsymbol{#1}}

\usepackage{adjustbox}
\usepackage{array}
\usepackage{booktabs}

\newcolumntype{R}[2]{%
    >{\adjustbox{angle=#1,lap=\width-(#2)}\bgroup}%
    l%
    <{\egroup}%
}

\usepackage[]{trackchanges}
\addeditor{YJ}
\addeditor{WD}
\addeditor{HC}

\title{Self-training with Two-phase Self-augmentation for Few-shot Dialogue Generation}

\author{Wanyu Du \quad Hanjie Chen \quad Yangfeng Ji \\
         Department of Computer Science \\ University of Virginia \\  
         Charlottesville, VA 22904\\
        \texttt{\{wd5jq,hc9mx,yangfeng\}@virginia.edu} \\
         }

\begin{document}
\maketitle

\begin{abstract}
In task-oriented dialogue systems, response generation from meaning representations (MRs) often suffers from limited training examples, due to the high cost of annotating MR-to-Text pairs.
Previous works on self-training leverage fine-tuned conversational models to automatically generate pseudo-labeled MR-to-Text pairs for further fine-tuning.
However, some self-augmented data may be noisy or uninformative for the model to learn from.
In this work, we propose a two-phase self-augmentation procedure to generate high-quality pseudo-labeled MR-to-Text pairs:
the first phase selects the most informative MRs based on model's prediction uncertainty; with the selected MRs, the second phase generates accurate responses by aggregating multiple perturbed latent representations from each MR. 
Empirical experiments on two benchmark datasets, \textsc{FewShotWOZ} and \textsc{FewShotSGD}, show that our method generally outperforms existing self-training methods on both automatic and human evaluations.\footnote{Please check the code, data, and evaluation scripts of this work at: \url{https://github.com/wyu-du/Self-Training-Dialogue-Generation}}
\end{abstract}

\section{Introduction}
\label{sec:intro}

\begin{figure*}[t]
  \includegraphics[width=\textwidth]{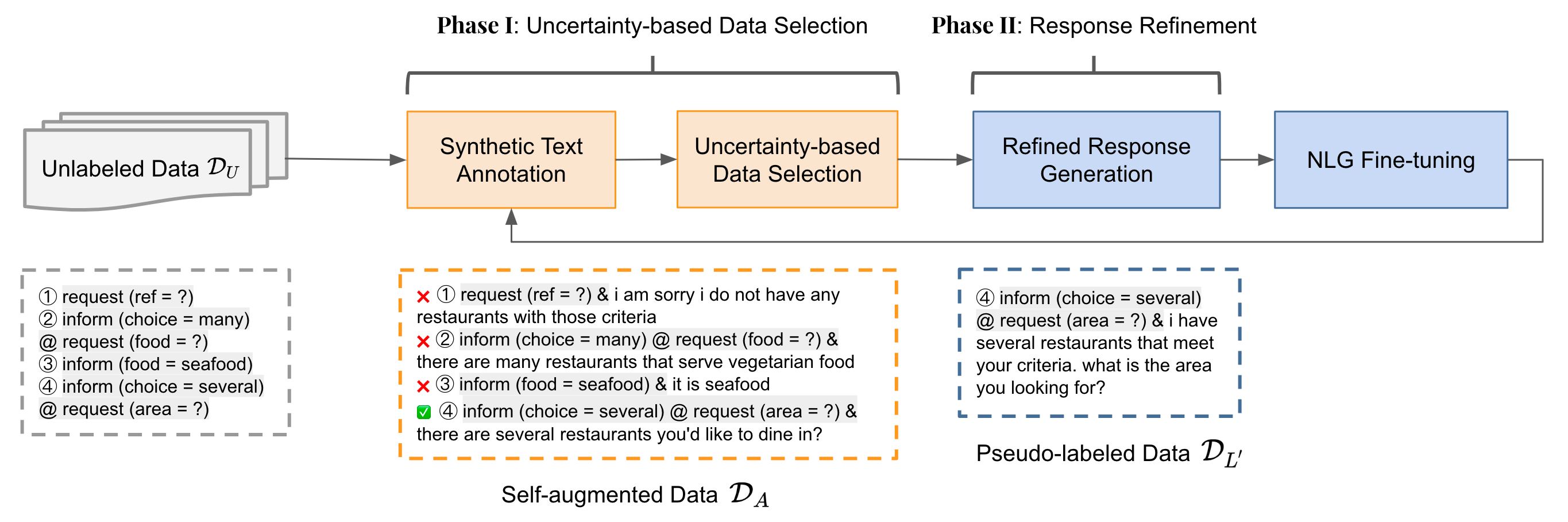}
  \caption{Our two-phase self-augmentation ($\textsc{SA}^2$) self-training framework for few-shot MR-to-Text generation. \label{fig:framework}}
\end{figure*}

\begin{table}[t]
  \centering
  \small
  \begin{tabular}{@{}l@{\hskip 2mm}p{0.32\textwidth}@{\hskip 1mm}c@{\hskip 1mm}c@{}}
    \toprule
     & \textbf{Self-augmented Data} & $\mathbb{E}[p_\theta]$ & $Var[p_\theta]$ \\
    \midrule
     1 & \colorbox{gray!20}{request (ref = ?) \&} i am sorry i do not have any restaurants with those criteria & \textcolor{red}{\textbf{low}} & \textcolor{red}{\textbf{low}} \\
     \midrule
     2 & \colorbox{gray!20}{inform (choice = many) @ request (foo}
     \colorbox{gray!20}{d = ?) \&} there are many restaurants that serve vegetarian food & \textcolor{red}{\textbf{low}} & \textcolor{teal}{\textbf{high}} \\
     \midrule
     3 & \colorbox{gray!20}{inform (food = seafood) \&} it is seafood & \textcolor{teal}{\textbf{high}} & \textcolor{red}{\textbf{low}} \\
     \midrule
     $\underset{\text{\cmark}}{4}$ & \colorbox{gray!20}{inform (choice = several) @ request (a}
     \colorbox{gray!20}{rea = ?) \&} there are several restaurants you'd like to dine in?
      & \underline{\textcolor{teal}{\textbf{high}}} & \underline{\textcolor{teal}{\textbf{high}}}\\
    \bottomrule
  \end{tabular}
  \caption{\label{tab:intro_example}
  Examples of our self-augmented data and data selection strategy.
  \colorbox{gray!20}{text} is the input MR (e.g.  \textit{request} is the dialogue intent, and \textit{(ref = ?)} is the slot-value pair of the current intent).
  The model $p_\theta$ generates synthetic dialogue response conditioning on the \colorbox{gray!20}{text}. 
  For each self-augmented data, a \textcolor{red}{\textbf{low}} predictive mean $\mathbb{E}[p_\theta]$ indicates that the model finds the augmented data ``too noisy'' (e.g. out-of-domain or invalid response), and a \textcolor{red}{\textbf{low}} predictive variance $Var[p_\theta]$ indicates that the model finds the augmented data ``too certain'' (e.g. uninformative response).
  In this work, we propose to select examples with \textcolor{teal}{\textbf{high}} $\mathbb{E}[p_\theta]$ and \textcolor{teal}{\textbf{high}} $Var[p_\theta]$.
  }
\end{table}

In task-oriented dialogue systems, a natural language generation (NLG) module is an essential component: it maps structured dialogue meaning representations (MRs) into natural language responses.
The NLG module has a great impact on users' experience because it directly interacts with users using text responses \citep{wen-etal-2015-semantically,Rastogi_Zang_Sunkara_Gupta_Khaitan_2020,kale-rastogi-2020-template,peng-etal-2020-shot}.
However, in real-world applications, developers often only have a few well-annotated data and confront a high data collection cost in specific domains.
This real-world challenge makes building an NLG module in the low-data setting a valuable research problem \citep{kale-rastogi-2020-template,chen-etal-2020-shot,peng-etal-2020-shot}.

While language models have been widely adopted to build the NLG module in task-oriented dialogue systems, they usually require thousands of MR-to-Text pairs for learning the domain-specific knowledge \citep{wen-etal-2016-multi,zhu-etal-2019-multi,Yang_Li_Quan_2021,lee-2021-improving-end}.
To collect more training data under a feasible budget, previous works propose three general approaches: (1) designing handcraft rules to augment new data, which is hard to scale up \citep{wei-zou-2019-eda,feng-etal-2020-genaug}; (2) building task-specific data retriever to search related data, which may overfit on the few training data \citep{xu-etal-2021-augnlg}; or (3) leveraging pre-trained language models to generate new data, which may generate ``too noisy'' data \citep{peng21b_interspeech,fabbri-etal-2021-improving,heidari-etal-2021-getting}.

Ideally, the augmented data should help the model better learn the domain-specific knowledge.
However, some augmented data can be ``too noisy'' that leads the model to learn irrelevant or inappropriate data patterns.
This phenomenon is also described as negative transfer in other works \citep{NIPS2011_93fb9d4b,Wang_2019_CVPR,meftah-etal-2021-hidden,feng-etal-2021-wasserstein}. 
To address this challenge, some works leverage human judgements to filter out the ``too noisy'' augmented data, which are difficult to scale up across different domains and tasks \citep{peris-casacuberta-2018-active,p-v-s-meyer-2019-data}.
Other works train task-specific discriminators to pick up the valid augmented data, which are likely to overfit in the low-data setting \citep{mi-etal-2021-self,xu-etal-2021-augnlg,bakshi-etal-2021-structure,heidari-etal-2021-getting,mehta2022improving}.

In this work, we propose to address the issue of selecting high-quality self-augmented examples with a two-phase procedure, where each phase will take care of selecting inputs and generating outputs independently.
As illustrated in \autoref{fig:framework}, the first phase evaluates input MRs with model's prediction uncertainty, aiming at selecting input examples that are informative to the current model. 
Specifically, for each input MR, we let the current model generate a response, and then apply the Monte Carlo Dropout method \citep{pmlr-v48-gal16} to estimate the predictive mean $\mathbb{E}[p_\theta]$ and predictive variance $Var[p_\theta]$ of the generated response.
In uncertainty quantification \citep{Gal2016Uncertainty}, high predictive mean indicates that the model is familiar with this input (i.e. in-domain data) and high predictive variance reflects that the model is sensitive to this input (i.e. informative data).
Hence, we propose to select input MRs with high predictive mean and variance. 
Note that our uncertainty-based data selection strategy neither requires training additional neural models to select the valid data \citep{bakshi-etal-2021-structure,heidari-etal-2021-getting,mehta2022improving}, nor need to calculate the data statistics across all training epochs and re-train the model overall again \citep{swayamdipta-etal-2020-dataset}.
The second phase aims at further improving the quality of the selected data. We adopt an idea from contrastive representation learning \citep{gao-etal-2021-simcse} and use the aggregation of randomly perturbed latent representations to help the model produce more accurate responses.
The combination of these two phases guarantees the proposed method selects more informative MR inputs and generates less noisy responses for further model fine-tuning.

In summary, the contributions of this work are as follows:
\begin{enumerate}
    \itemsep -0.3em
    \item Proposing a novel self-training algorithm for the few-shot MR-to-Text generation problem in task-oriented dialogue systems, which applies a two-phase self-augmentation strategy to identify informative MRs and generate accurate responses for further fine-tuning.
    \item Showing that the proposed method generally outperforms other few-shot NLG baselines on two benchmark datasets, \textsc{FewShotWOZ} \citep{peng-etal-2020-shot} and \textsc{FewShotSGD} \citep{xu-etal-2021-augnlg} in both automatic and human evaluations.
    \item Conducting in-depth empirical analysis on key components of the proposed few-shot self-training framework: the pre-trained language model, the data selection strategy, and the model training configurations.
\end{enumerate}

\section{Related Works}

\paragraph{Task-oriented dialogue generation.}
Previous NLG methods generate system responses by: (1) designing handcraft response templates and filling in slot-value pairs from system actions, or (2) building data-driven neural models, which encode systems actions into latent feature representations and decode natural language responses with more diversity in realization.  
However, both approaches cause high data collection costs.
The template-based methods \citep{langkilde-knight-1998-generation-exploits,Cheyer:99043} require collecting a comprehensive set of templates to cover all possible combinations of dialog acts and slot-value pairs,
while data-driven methods \citep{wen-etal-2015-semantically,wen-etal-2017-network,zhu-etal-2019-multi} require collecting thousands of system action and response pairs to ensure the neural model generating fluent responses.

\paragraph{Few-shot NLG.}
Recent works on few-shot NLG mainly focus on developing or adapting pre-trained language models. 
\citet{peng-etal-2020-shot} presents the first few-shot NLG benchmark for task-oriented dialog systems, and develops a pre-trained language model which can be fine-tuned with only a few domain-specific labels to adapt to new domains.
\citet{chen-etal-2020-shot} applies the switch mechanism to combine the information from both input data and pre-trained language models, which achieves good performance in table-to-text generation tasks.
\citet{chang-etal-2021-training} studies the training data selection strategies in few-shot NLG, and finds that clustering-based selection strategy consistently helps generative models get better performance than randomly sampling.

\paragraph{Self-training for NLG.}
There has been some works applying the self-training technique to improve the model's generalization ability in NLG tasks.
Some works \citep{mi-etal-2021-self,xu-etal-2021-augnlg} leverage the self-training framework to pseudo-label the unlabeled data and select the training data based on the confidence score from a single student model.
Other works \citep{kedzie-mckeown-2019-good,He2020Revisiting} show that the noisy self-training is able to utilize unlabeled data and improve the performance of the supervised baseline.
However, their observations come from large-scale training datasets, which may not necessarily hold in the few-shot data setting, because a single Transformer-based model may heavily overfit on the few-shot training data in the early iteration.

We also find some works \citep{bakshi-etal-2021-structure,heidari-etal-2021-getting,mehta2022improving} leverage generation models to produce pseudo-labeled data. 
However, they train additional neural models to select the pseudo-labeled data.
\citet{bakshi-etal-2021-structure} and \citet{heidari-etal-2021-getting} use the reconstruction loss from a fine-tuned BART model \citep{lewis-etal-2020-bart} to select the pseudo-labeled data.
Besides, \citet{mehta2022improving} leverage a fine-tuned BLEURT model \citep{sellam-etal-2020-bleurt} with a selection threshold to select pseudo-responses for self-training.
Intuitively, the pseudo-labeled data should bring new domain-specific knowledge to the model.
While prior works select the pseudo-labeled data using an independent neural model, we propose to select the pseudo-labeled data using the generation model itself and eliminate the requirement for training additional models.

\paragraph{Data selection strategies.}
Some works in active learning leverage human judgments to select the augmented data.
\citet{peris-casacuberta-2018-active,p-v-s-meyer-2019-data} design data selection functions to select a subset of representative unlabeled data for humans to annotate, and get better model performance by leveraging human annotation.
However, the additional requirement of human judgments will increase the difficulty of adapting the method across different domains.
Another work \citep{swayamdipta-etal-2020-dataset} leverages the model training dynamics to categorize and select the data, but their method requires massive ground-truth labeled data.
In contrast, our self-training framework does not require additional human judgments or massive ground-truth labeled data, which can be easily adapted to different tasks across different domains.
\section{Proposed Method}

In task-oriented dialogue systems, the NLG module translates a structured dialogue meaning representation $\mathcal{A}$ into a natural language response $\vec{x}=\{x_1, ..., x_T\}$.
One structured dialogue meaning representation $\mathcal{A}$ consists of $K$ dialogue intents 
and a list of slot-value pairs for each intent:
\begin{equation}
    \mathcal{A} = \{\mathcal{I}_k, (s_{k,1}, v_{k,1}), ..., (s_{k,P_k}, v_{k,P_k})\}_{k=1}^K
\end{equation}
where the dialogue intent $\mathcal{I}_k$ indicates different types of system actions and the slot-value pairs $\{(s_{k,i}, v_{k,i})\}_{i=1}^{P_{k}}$ shows the category names and their content information to be expressed in the response.
For example, \emph{inform (area = west; choice = many)}, where \emph{inform} is the dialogue intent, \emph{area} and \emph{choice} are the slot names, \emph{west} and \emph{many} are the slot values.

We define $p_\theta(\vec{x}\mid \mathcal{A})$ as the generation model that generates the response $\vec{x}$ in an auto-regressive way conditioning on $\mathcal{A}$:
\begin{equation}
\label{eq:ori_p}
    p_\theta(\vec{x}\mid \mathcal{A}) = \prod_{t=1}^T p_\theta(x_t \mid x_{1:t-1}, \mathcal{A})
\end{equation}
where $\theta$ is the model parameter.
{A typical way of learning $\theta$ is by} maximizing the log-likelihood of the conditional probabilities in \autoref{eq:ori_p} over the original training set $\mathcal{D}_L$:
\begin{equation}
    \mathcal{L}_\theta(\mathcal{D}_L) = \sum_{n=1}^{|D_L|}\sum_{t=1}^{T_n} \log{p_\theta(x_{t,n} \mid x_{1:t-1, n}, \mathcal{A}_n)}
\end{equation}
In the few-shot MR-to-Text generation setting, the size of training data $|D_L|$ is a small number (e.g. $\leq 50$).

\subsection{Self-training with Two-phase Self-augmentation ($\textsc{SA}^2$)}
The $\textsc{SA}^2$ self-training algorithm starts from a warm-up stage, where a base generation model is trained on the original training set $\mathcal{D}_L$ for a few epochs. 
Then, in each iteration of self-training, the algorithm consists of four steps: synthetic text annotation, uncertainty-based data selection, response refinement, and model fine-tuning. 

The synthetic text annotation uses the current model to generate synthetic text responses based on input MRs and constructs a preliminary version of self-augmented data $\mathcal{D}_A$.
Next, the data selection uses the prediction uncertainty of the current model on the synthetic responses to select informative MRs in $\mathcal{D}_A$, which is the \emph{first phase} of self-augmentation.
Given the selected MRs, the \emph{second phase} of self-augmentation is to generate more accurate text responses via aggregating multiple latent representations from model parameters with different dropout masks, which produces the pseudo-labeled data $\mathcal{D}_{L'}$.
Finally, the current model is fine-tuned with both the original training set $\mathcal{D}_{L}$ and the pseudo-labeled dataset $\mathcal{D}_{L'}$.

The detailed procedure of $\textsc{SA}^2$ self-training algorithm is demonstrated in \autoref{alg:alg}.
We describe the proposed uncertainty-based data selection method in \S\ref{sec:data_selection} and response refinement method in \S\ref{sec:response_generation} respectively.

\begin{algorithm}[ht]
  \caption{$\textsc{SA}^2$ Self-training Algorithm}
  \label{alg:alg}
  \textbf{Input}: The original training set $\mathcal{D}_L$, in-domain MRs $\mathcal{D}_U$, base generation model $p_\theta$, number of self-training iterations $S$ \\
  \textbf{Output}: A fine-tuned generation model $p_\theta$ \vspace{-4mm}
  \begin{algorithmic}[1]
    \STATE Load $p_\theta$ and train $p_\theta$ on $\mathcal{D}_L$
    \FOR {$s = 1,\dots,S$}
    \STATE Initialize $\mathcal{D}_{A}=\emptyset$ and $\mathcal{D}_{L'}=\emptyset$
    \STATE \textit{// Synthetic Text Annotation}
    \FOR {$\mathcal{A}_n \in \mathcal{D}_U$}
    \STATE Generate $\vec{x}_n \sim p_\theta(\vec{x}_n \mid \mathcal{A}_n)$
    \STATE $\mathcal{D}_{A} \cup \{(\vec{x}_n, \mathcal{A}_n)\}$
    \ENDFOR
    \STATE \textit{// Data Selection}
    \STATE Compute threshold $\bar{\mu}$ and $\bar{s}$ using Eq. (\ref{eq:threshold})
    \FOR {$(\vec{x}_n, \mathcal{A}_n) \in \mathcal{D}_{A}$}
    \IF{$\mathbb{E}[p_\theta] > \bar{\mu}$ and $Var[p_\theta] > \bar{s}$}
    \STATE \textit{// Response Refinement}
    \STATE Generate $\bar{\vec{x}}_n$ using Eq.(\ref{eq:h_bar})
    \STATE $\mathcal{D}_{L'} \cup \{(\bar{\vec{x}}_n, \mathcal{A}_n)\}$
    \ENDIF
    \ENDFOR
    \STATE Fine-tune $p_\theta$ on $\mathcal{D}_{L} \cup \mathcal{D}_{L'}$
    \ENDFOR
  \end{algorithmic}
\end{algorithm}
\subsection{Phase I: Uncertainty-based Data Selection}
\label{sec:data_selection}
We hypothesize that the generation model is likely to gain little by learning from the data, if
(1) it finds ``too noisy'', which may be out-of-domain or invalid; 
(2) it finds ``too certain'', which may be uninformative to learn from.
Therefore, we propose to select the data which the current model finds ``less noisy'' and ``more uncertain''.
Intuitively, data with ``less noise'' may provide helpful domain-specific knowledge to the model, meanwhile ``more uncertainty'' indicates the model has not learned well from the data yet, thus may produce incoherent responses.

\paragraph{Uncertainty estimation.}
We use the Monte Carlo Dropout method \citep{pmlr-v48-gal16,NEURIPS2020_f23d125d} to estimate the ``noise'' and ``uncertainty'' of each self-augmented data regarding the current model.
For each self-augmented data $(\vec{x}, \mathcal{A})$, we enable dropouts before every hidden layer in the generation model, perform $M$ forward passes through the model, and get $M$ i.i.d. model likelihood estimations $\{p_{\theta_i}(\vec{x}\mid \mathcal{A})\}_{i=1}^M$. 
These $M$ outputs are empirical samples of an approximated posterior distribution $p(\vec{x}\mid \mathcal{A})$ \citep{Gal2016Uncertainty}.
Then, we compute the predictive mean $\mathbb{E}[p_\theta]$ of the approximated distribution $p(\vec{x}\mid \mathcal{A})$ and predictive variance $Var[p_\theta]$ of the empirical samples:
\begin{eqnarray}
    \hspace{-3mm} \mathbb{E}[p_\theta] &\approx& \frac{1}{M}\sum_{i=1}^M p_{\theta_i}(\vec{x}\mid \mathcal{A}) \label{eq:mean} \\
    \hspace{-3mm} Var[p_\theta] &\approx& \frac{1}{M}\sum_{i=1}^M (p_{\theta_i}(\vec{x}\mid \mathcal{A}) - \mathbb{E}[p_\theta])^2 \label{eq:var}
\end{eqnarray}
A low predictive mean $\mathbb{E}[p_\theta]$ means the model finds the current data ``too noisy'', because it has a low likelihood estimation of the current data, which indicates the current data may be out-of-domain or invalid;
while a low predictive variance $Var[p_\theta]$ means the model finds the current data ``too certain'', because all empirical samples have a similar likelihood estimation of the current data, which indicates the current data may be uninformative for the model to learn from.
Therefore, we consider self-augmented data with both high predictive means and variances are examples of interest.

\paragraph{Selection strategy.}
The next question is \emph{what are the thresholds for high predictive means and variances?}
First, we calculate the corpus-level predictive mean $\mu_A$ of the self-augmented $\mathcal{D}_A$, and filter out the augmented data which have a lower predictive mean than $\mu_A$, because we observe that such data are often very noisy and contain many redundant slots. 
Then, we combine and sort the original training data $\mathcal{D}_L$ and the remaining self-augmented data, and further remove the outliers (i.e. first and last 1\% of datapoints).
Assume that the collection of predictive mean scores $\mathbb{E}[p_\theta]$ and variance scores $Var[p_\theta]$ of the selected data follows a Gaussian distribution respectively, then the data selection threshold is defined as
\begin{equation}
\label{eq:threshold}
    \bar{\mu} = \frac{1}{N}\sum_{n=1}^{N} p_n, ~~~
    \bar{s} = \frac{1}{N}\sum_{n=1}^{N} v_n
\end{equation}
where $p_n$ is the predictive mean and $v_n$ is the predictive variance of the $n$-th selected data, $N$ is the total number of original training data and remaining self-augmented data (after removing the outliers).

We select the self-augmented data with high $\mathbb{E}[p_\theta]$ (above the average predictive mean $\bar{\mu}$) and high $Var[p_\theta]$ (above the average predictive variance $\bar{s}$).
We also explored other data selection strategies (detailed in \S\ref{sec:other_analysis}),
and find that selecting high $\mathbb{E}[p_\theta]$ and high $Var[p_\theta]$ data empirically brings more performance improvements than other strategies.

\subsection{Phase II: Response Refinement}
\label{sec:response_generation}
Since the large generation model is trained on a small training set, it is very likely to overfit and produce high-biased latent representations that cause the generation of {inaccurate} text responses.
To reduce the risk of producing high-biased latent representations, we adopt dropout noise proposed in contrastive learning \citep{gao-etal-2021-simcse} into the latent representation during inference.   

Specifically, {for each selected input MR from \textbf{Phase I}}, we enable the dropout masks of the model (placed on fully-connected layers as well as attention probabilities) at the decoding timestamp $t$, and compute $R$ latent representations {$\{\vec{h}_{\theta_i}^t\}_{i=1}^{R}$}, then take an average over all latent representations to obtain the final latent representation for the current probability distribution:
\begin{equation}
\label{eq:h_bar}
    p(\bar{x}_t \mid \bar{x}_{1:t-1}, \mathcal{A}) = \text{softmax}(\frac{1}{R} \sum_{i=1}^R \vec{h}_{\theta_i}^t) 
\end{equation}
Then, we generate the text response $\bar{\vec{x}}$ according to the probability distribution $p(\bar{x}_t \mid \bar{x}_{1:t-1}, \mathcal{A})$ and add the data $(\bar{\vec{x}}, \mathcal{A})$ into the pseudo-labeled dataset $\mathcal{D}_{L'}$.
We fine-tune the generation model on both the original training set $\mathcal{D}_{L}$ and the pseudo-labeled dataset $\mathcal{D}_{L'}$.
Fine-tuning the refined responses is shown to improve the model's final performances (detailed in \S\ref{sec:ablation}).
\section{Experiments}

\begin{table*}[t]
  \centering
  \footnotesize
  \begin{tabular}{l@{\hskip 1mm}|r@{\hskip 1mm}r@{\hskip 1mm}|r@{\hskip 1mm}r@{\hskip 1mm}|r@{\hskip 1mm}r@{\hskip 1mm}|r@{\hskip 1mm}r@{\hskip 1mm}|r@{\hskip 1mm}r@{\hskip 1mm}|r@{\hskip 1mm}r@{\hskip 1mm}|r@{\hskip 1mm}r@{\hskip 1mm}}
    \toprule
     & \multicolumn{2}{c|}{\textbf{Restaurant}} & \multicolumn{2}{c|}{\textbf{Laptop}} & \multicolumn{2}{c|}{\textbf{Hotel}} & \multicolumn{2}{c|}{\textbf{TV}} & \multicolumn{2}{c|}{\textbf{Attraction}} & \multicolumn{2}{c|}{\textbf{Train}} & \multicolumn{2}{c}{\textbf{Taxi}} \\
     & BLEU & ERR & BLEU & ERR & BLEU & ERR & BLEU & ERR & BLEU & ERR & BLEU & ERR & BLEU & ERR \\
    \midrule
    \textbf{SC-GPT} & 34.62 & \textbf{1.95} & 33.31 & 3.01 & 40.74 & 3.55 & 33.72 & 1.72 & 23.77 & 1.40 & 25.09 & 1.90 & 18.22 & 0.00 \\
    \textbf{AUG-NLG} & 29.94 & 2.28 & 30.02 & 4.29 & 38.30 & 4.73 & 32.41 & 3.34 & 21.76 & 3.95 & 24.06 & 3.81 & 17.99 & 0.00 \\
    \midrule
    \textbf{ST-ALL} & 33.84 & 6.51 & 34.40 & 4.28 & 39.68 & 1.78 & 34.88 & 1.76 & 24.32 & 3.19 & 24.47 & 3.87 & 17.89 & 0.00 \\
    \textbf{ST-NLL} & 33.07 & 9.44 & 34.99 & 3.37 & 41.40 & 5.92 & 35.98 & 2.26 & 24.87 & 4.85 & 23.53 & 5.27 & 20.21 & 0.00 \\
    \textbf{ST-$\textsc{SA}^2$} (ours) & \textbf{36.48} & 2.60 & \textbf{35.42} & \textbf{2.04} & \textbf{42.63} & \textbf{1.77} & \textbf{36.39} & \textbf{1.63} & \textbf{25.63} & \textbf{1.40} & \textbf{25.34} & \textbf{1.62} & \textbf{20.95} & \textbf{0.00} \\
    \bottomrule
  \end{tabular}
  \caption{\label{tab:bleu_woz}
  Automatic evaluation results on the test set of \textsc{FewShotWOZ} (BLEU$\uparrow$, ERR$\downarrow$). The results of \textbf{AUG-NLG} come from the data and code released by \citet{xu-etal-2021-augnlg}, the other results come from our implementation.
  }
\end{table*}

\begin{table*}[t]
  \centering
  \small
  \begin{tabular}{l|rrrrrrrr}
    \toprule
     & \textbf{Restaurants} & \textbf{Hotels} & \textbf{Flights} & \textbf{Buses} & \textbf{Events} & \textbf{Rentalcars} & \textbf{Services} & \textbf{Ridesharing} \\
    \midrule
    \textbf{SC-GPT} & 19.86 & 22.21 & 26.63 & 19.87 & \textbf{26.41} & 20.21 & 27.32 & 22.03 \\
    \textbf{AUG-NLG} & 19.73 & 12.38 & 23.20 & 16.81 & 19.62 & 16.64 & 20.18 & 17.20 \\
    \midrule
    \textbf{ST-ALL} & 19.71 & 21.45 & 26.90 & 19.76 & 25.68 & 20.22 & 27.59 & 21.14 \\
    \textbf{ST-NLL} & 14.52 & 21.29 & \textbf{27.59} & 20.27 & 25.81 & 20.07 & 26.54 & 19.84 \\
    \textbf{ST-$\textsc{SA}^2$} (ours) & \textbf{20.42} & \textbf{22.90} & 27.12 & \textbf{21.16} & 25.32 & \textbf{20.70} & \textbf{28.34} & \textbf{23.28} \\
    \bottomrule
    \toprule
    & \textbf{Movies} & \textbf{Calendar} & \textbf{Banks} & \textbf{Music} & \textbf{Homes} & \textbf{Media} & \textbf{Travel} & \textbf{Weather} \\
    \midrule
    \textbf{SC-GPT} & 25.71 & 23.53 & 25.99 & 24.01 & 24.90 & 26.24 & 24.97 & 27.89 \\
    \textbf{AUG-NLG} & 16.93 & 13.60 & 12.89 & 9.56 & 18.06 & 10.51 & 15.77 & 10.74 \\
    \midrule
    \textbf{ST-ALL} & 26.19 & 24.86 & 25.03 & 24.62 & 24.97 & 26.56 & 25.28 & 28.06 \\
    \textbf{ST-NLL} & 23.98 & 23.67 & 25.70 & 18.88 & 24.82 & 26.99 & 24.95 & 28.64 \\
    \textbf{ST-$\textsc{SA}^2$} (ours) & \textbf{28.95} & \textbf{25.24} & \textbf{28.14} & \textbf{27.23} & \textbf{25.03} & \textbf{28.76} & \textbf{25.34} & \textbf{29.27} \\
    \bottomrule
  \end{tabular}
  \caption{\label{tab:bleu_sgd}
  Automatic evaluation results of BLEU scores on the test set of \textsc{FewShotSGD}. The results of \textbf{AUG-NLG} come from the data and code released by \citet{xu-etal-2021-augnlg}, the other results come from our implementation.
  }
\end{table*}

We conduct experiments to answer three research questions:
(1) Is $\textsc{SA}^2$ self-training algorithm a helpful method to deal with the few-shot dialogue generation problem?
(2) Can our data selection strategy effectively filter out the ``too noisy'' and ``uninformative'' augmented data?
(3) Can our response refinement method help improve the performance of the NLG model?

\subsection{Setups}
\paragraph{Benchmark datasets.}
We evaluate our method on two few-shot dialogue generation benchmark datasets: \textsc{FewShotWOZ} \citep{peng-etal-2020-shot} and \textsc{FewShotSGD} \citep{xu-etal-2021-augnlg}.
\textsc{FewShotWOZ} has 7 domains and an average number of 50 training examples per domain.
\textsc{FewShotSGD} has 16 domains and an average number of 35 training examples per domain.
However, both datasets do not provide the development sets for hyper-parameter tuning.
To create the standard training/dev/test data splits, we randomly sampled 10\% data from the original test set as the dev set, and kept the training set unchanged. 
For fair comparisons across different methods, we evaluated all methods on the new split test set.
The detailed data statistics of the two benchmarks are described in \autoref{sec:appendix-dataset}.  

\paragraph{Unlabeled data.}
The two benchmark datasets are sampled and constructed based on the three datasets: RNNLG \citep{wen-etal-2016-multi}, MultiWOZ \citep{budzianowski-etal-2018-multiwoz} and SGD \citep{rastogi2020towards}.
To ensure the input MRs are within the same domain of the original training set $\mathcal{D}_L$, we collect all augmented MRs from the training set of RNNLG, MultiWOZ, and SGD.
For \textsc{FewShotWOZ}, we collect an average number of 9,080 unlabeled MRs per domain.
For \textsc{FewShotSGD}, we collect an average number of 7,532 unlabeled MRs per domain.
The detailed data statistics of each domain are demonstrated in \autoref{sec:appendix-dataset}.

\paragraph{Baselines.}
We compare our method with four baselines and describe the model configuration and training details in \autoref{sec:appendix-implementation}.
(1) \textbf{SC-GPT} \citep{peng-etal-2020-shot} is the state-of-the-art pre-trained language model for NLG in task-oriented dialogue systems, which is further fine-tuned on each specific domain using the original training data $\mathcal{D}_L$;
(2) \textbf{AUG-NLG} \citep{xu-etal-2021-augnlg} leverages the pre-trained SC-GPT model, first trains it on its automatically retrieved augmented data, then fine-tunes it on each few-shot domain;
(3) \textbf{ST-ALL} is the traditional self-training baseline which learns from all self-augmented data without any data selection and text refinement;
(4) \textbf{ST-NLL} adopts the traditional self-training baseline but learns from the self-augmented data which has a lower than the average reconstruction loss according to the current generation model;
(5) \textbf{ST-$\textsc{SA}^2$} is our method, in addition to our proposed data selection strategy and response refinement method, we apply a rule-based parser \citep{kedzie-mckeown-2019-good} to heuristically filter out invalid responses that do not match the slot-value pairs in the input MRs on the \textsc{FewShotWOZ} dataset in order to achieve lower ERR.

\begin{table*}[t]
  \centering
  \footnotesize
  \begin{tabular}{l@{\hskip 1mm}|r@{\hskip 1mm}r@{\hskip 1mm}|r@{\hskip 1mm}r@{\hskip 1mm}|r@{\hskip 1mm}r@{\hskip 1mm}|r@{\hskip 1mm}r@{\hskip 1mm}|r@{\hskip 1mm}r@{\hskip 1mm}|r@{\hskip 1mm}r@{\hskip 1mm}|r@{\hskip 1mm}r@{\hskip 1mm}}
    \toprule
     & \multicolumn{2}{c|}{\textbf{Restaurant}} & \multicolumn{2}{c|}{\textbf{Laptop}} & \multicolumn{2}{c|}{\textbf{Hotel}} & \multicolumn{2}{c|}{\textbf{TV}} & \multicolumn{2}{c|}{\textbf{Attraction}} & \multicolumn{2}{c|}{\textbf{Train}} & \multicolumn{2}{c}{\textbf{Taxi}} \\
     & BLEU & ERR & BLEU & ERR & BLEU & ERR & BLEU & ERR & BLEU & ERR & BLEU & ERR & BLEU & ERR \\
    \midrule
    \textbf{ST-$\textsc{SA}^2$} (ours) & \textbf{36.48} & \textbf{2.60} & \textbf{35.42} & \textbf{2.04} & \textbf{42.63} & \textbf{1.77} & \textbf{36.39} & \textbf{1.63} & \textbf{25.63} & \textbf{1.40} & \textbf{25.34} & \textbf{1.62} & \textbf{20.95} & \textbf{0.00} \\
    \textbf{~~w/o aggregation} & 35.30 & 3.25 & 34.30 & 3.57 & 39.08 & 2.96 & 36.24 & 5.25 & 24.44 & 2.55 & 24.15 & 2.01 & 20.17 & 1.69 \\
    \textbf{~~w/o filter} & 36.17 & 3.90 & 34.19 & 5.85 & 39.52 & 3.55 & 35.45 & 2.76 & 25.51 & 2.42 & 24.89 & 2.24 & 20.60 & 0.00 \\
    \bottomrule
  \end{tabular}
  \caption{\label{tab:ablation_woz}
  Ablation study results on the test set of \textsc{FewShotWOZ} (BLEU$\uparrow$, ERR$\downarrow$).
  }
\end{table*}
\begin{table}[t]
  \centering
  \small
  \begin{tabular}{lcc}
    \toprule
     & \textbf{Informativeness} $\uparrow$ & \textbf{Naturalness} $\uparrow$ \\
    \midrule
    \textbf{SC-GPT} & 2.62 & 2.32 \\
    \textbf{ST-NLL} & 2.69 & 2.31 \\ 
    \textbf{ST-$\textsc{SA}^2$} (ours) & \textbf{2.69} & \textbf{2.41} \\
    \midrule
    \textit{Human} & 2.71 & 2.49 \\
    \bottomrule
  \end{tabular}
  \caption{\label{tab:human_eval_woz}
  Human evaluation results on the sampled test set of \textsc{FewShotWOZ}.
  }
\end{table}
\paragraph{Automatic evaluation.}
We follow the prior works \citep{wen-etal-2015-semantically,peng-etal-2020-shot,xu-etal-2021-augnlg} and use BLEU score and Slot Error Rate (ERR) for automatic evaluation.
ERR is computed by exact matching the slot tokens in the generated responses as $ERR=(p+q)/N$, where $N$ is the total number of slots in the MR, and p,q is the number of missing and redundant slots in the generated response.
For each MR, we generate five responses and select the top one with the lowest ERR as the final output.
Note that we only compute ERR on the \textsc{FewShotWOZ} dataset, because the \textsc{FewShotSGD} dataset does not release its evaluation script.

\paragraph{Human evaluation.}
We follow the prior works \citep{peng-etal-2020-shot,kale-rastogi-2020-template} and use Amazon Mechanical Turk to conduct human evaluation.
We recruited master level workers with over 90\% approval rate to compare and rate the responses generated by different methods and the the ground truth response.
The workers are asked to rate the response on a scale of 1 (bad) to 3 (good) in terms of \textit{informativeness} and \textit{naturalness}.
Informativeness indicates how much information from the input MR has been covered in the response, and naturalness measures whether the response looks coherent, grammatical, and natural.
Each data pair is rated by 3 workers.
We randomly sample 120 examples from each dataset, and collect a total of 2880 ratings.

\subsection{Result Analysis}
\paragraph{On \textsc{FewShotWOZ}.}
The automatic evaluation results in \autoref{tab:bleu_woz} show that \textbf{ST-$\textsc{SA}^2$} outperforms other baselines across all domains in both BLEU and ERR.
Besides, we observe that \textbf{SC-GPT} is a strong baseline, and \textbf{ST-NLL} can bring more performance improvements than \textbf{AUG-NLG} and \textbf{ST-ALL} in 5 out of 7 domains, which shows the effectiveness of data selection in self-training.
The human evaluation results in \autoref{tab:human_eval_woz} indicate that \textbf{ST-$\textsc{SA}^2$} can generate more natural and informative responses than \textbf{SC-GPT} and \textbf{ST-NLL}.
We provide some model generation results of different methods in \autoref{sec:appendix-generation}.

\paragraph{On \textsc{FewShotSGD}.}
The automatic evaluation results in
\autoref{tab:bleu_sgd} illustrate that \textbf{ST-$\textsc{SA}^2$} outperforms other baselines in 14 out of 16 domains in BLEU score.
Additionally, we find that \textbf{ST-ALL} generally outperforms \textbf{AUG-NLG}, which indicates that additional pre-training on the retrieved task-relevant data does not necessarily help the model generate better responses.
In contrast, the self-training method \textbf{ST-ALL} generally improves the model performances in 10 out of 16 domains, which shows the benefit of learning from self-augmented data.
The human evaluation results in \autoref{tab:human_eval_sgd} demonstrate that \textbf{ST-$\textsc{SA}^2$} is capable to generate more informative and natural responses than \textbf{SC-GPT} and \textbf{ST-ALL}.
We provide some model generation results of different methods in \autoref{sec:appendix-generation}.

\begin{table}[t]
  \centering
  \small
  \begin{tabular}{lcc}
    \toprule
     & \textbf{Informativeness} $\uparrow$ & \textbf{Naturalness} $\uparrow$ \\
    \midrule
    \textbf{SC-GPT} & 2.53 & 2.31 \\
    \textbf{ST-ALL} & 2.55 & 2.40 \\ 
    \textbf{ST-$\textsc{SA}^2$} (ours) & \textbf{2.69} & \textbf{2.42} \\
    \midrule
    \textit{Human} & 2.69 & 2.56 \\
    \bottomrule
  \end{tabular}
  \caption{\label{tab:human_eval_sgd}
  Human evaluation results on the sampled test set of \textsc{FewShotSGD}.
  }
\end{table}
\subsection{Ablation Study on Response Refinement}
\label{sec:ablation}
To validate the effectiveness of the proposed response refinement method, we conduct ablation study on \textbf{ST-$\textsc{SA}^2$} by removing the representation aggregation in \autoref{eq:h_bar} and the rule-based filter \citep{kedzie-mckeown-2019-good} respectively.
We observe from \autoref{tab:ablation_woz} that removing the representation aggregation during response refinement will lead to degraded performances in both BLEU and ERR across all domains, which indicates the importance of obtaining lower-biased latent representations during self-augmentation.
Besides, we find that removing the rule-based filter will lead to worse performances in ERR across all domains, which reveals that the model is still likely to generate incorrect responses, and those incorrect pseudo-labeled data will cause the model to learn irrelevant patterns and perform worse on the unseen test set.

\begin{table}[t]
  \centering
  \small
  \begin{tabular}{lcccc}
    \toprule
    & $\mathbb{E}[p_\theta]$ & $Var[p_\theta]$ & \textbf{BLEU} $\uparrow$ & \textbf{ERR} $\downarrow$ \\
    \midrule
    1 & \textcolor{red}{\textbf{low}} & \textcolor{red}{\textbf{low}} & 32.72 & 1.62 \\
    2 & \textcolor{red}{\textbf{low}} & \textcolor{teal}{\textbf{high}} & 32.24 & 1.62 \\
    3 & \textcolor{teal}{\textbf{high}} & \textcolor{red}{\textbf{low}} & 33.18 & 2.28 \\
    4 & \textcolor{teal}{\textbf{high}} & \textcolor{teal}{\textbf{high}} & \textbf{36.48} & 2.60 \\
    \bottomrule
  \end{tabular}
  \caption{\label{tab:selection}
  Different data selection strategy comparison of \textbf{ST-$\textsc{SA}^2$} in the \textbf{Restaurant} domain on the test set of \textsc{FewShotWOZ}.
  }
\end{table}
\begin{table}[t]
  \centering
  \small
  \begin{tabular}{llcc}
    \toprule
    & \textbf{Base Model} & \textbf{BLEU} $\uparrow$ & \textbf{ERR} $\downarrow$ \\
    \midrule
    1 & {GPT2}  & 24.22 & 13.68 \\
    2 & {DialoGPT}  & 14.77 & 20.84 \\
    3 & {SC-GPT}  & \textbf{36.48} & 2.60 \\
    \bottomrule
  \end{tabular}
  \caption{\label{tab:base_model}
  Different base generation model comparison of \textbf{ST-$\textsc{SA}^2$} in the \textbf{Restaurant} domain on the test set of \textsc{FewShotWOZ}.
  }
\end{table}
\begin{table}[t]
  \centering
  \small
  \begin{tabular}{l@{\hskip 2mm}c@{\hskip 2mm}c@{\hskip3mm}c@{\hskip 3mm}c@{\hskip 3mm}c@{\hskip 1mm}}
    \toprule
    & \textbf{Epoch} & \textbf{LR} & $\text{\textbf{BLEU}}_{dev}\uparrow$ & $\text{\textbf{BLEU}}_{test}\uparrow$ & $\text{\textbf{ERR}}_{test} \downarrow$ \\
    \midrule
    1 & 10 & 1e-6 & 23.22 & 24.75 & 2.93 \\
    2 & 20 & 1e-6 & 22.96 & 24.63 & 2.04 \\
    3 & 20 & 5e-7 & \textbf{23.43} & \textbf{25.63} & \textbf{1.40} \\
    4 & 20 & 5e-8 & 23.29 & 24.82 & 1.91 \\
    \bottomrule
  \end{tabular}
  \caption{\label{tab:hyper_param}
  Different training hyper-parameters comparison of \textbf{ST-$\textsc{SA}^2$} in the \textbf{Attraction} domain of \textsc{FewShotWOZ}, where \textbf{Epoch} is the number of training epochs within a self-training iteration, and \textbf{LR} is the initial learning rate at the beginning of each training epoch. We select the best model which has the highest $\text{\textbf{BLEU}}_{dev}$.
  }
\end{table}
\subsection{Analysis of Other Components in $\textsc{SA}^2$ Self-training Algorithm}
\label{sec:other_analysis}

In this section, we provide additional empirical analysis on other components that will affect the performance of the $\textsc{SA}^2$ self-training algorithm, in order to gain more insights about the self-training technique in solving the few-shot NLG problem.

\paragraph{Data selection strategies.}
\autoref{tab:selection} compares different data selection strategies of \textbf{ST-$\textsc{SA}^2$} in the restaurant domain of \textsc{FewShotWOZ}.
We find that selecting low $\mathbb{E}[p_\theta]$ data will lead to degraded performance in BLEU score, because low $\mathbb{E}[p_\theta]$ data often contains more redundant tokens compared with the ground-truth response.
Although low $\mathbb{E}[p_\theta]$ data gives lower ERR, the generated texts are not very natural and fluent.
Selecting high $\mathbb{E}[p_\theta]$ and low $Var[p_\theta]$ data will also lead to degraded performance in the BLEU score, which is probably because the model overfits on the uninformative data.
We provide some self-augmented and pseudo-labeled examples of different data selection strategies in \autoref{sec:appendix-augment}.

\paragraph{Base generation models.}
For the base generation model selection, we compare different pre-trained language models, including GPT2 \citep{radford2019language}, DialoGPT \citep{zhang-etal-2020-dialogpt} and SC-GPT.
GPT2 is an open-end text generation model, and DialoGPT is an open-domain dialogue generation model.
In contrast, SC-GPT is trained on around 400K MR-to-Text pairs in task-oriented dialogue generation datasets.
As can be seen in \autoref{tab:base_model}, SC-GPT gives much better performance than GPT2 and DialoGPT, which indicates that selecting a suitable base generation model is critical for self-training.

\paragraph{Training hyper-parameters.}
\autoref{tab:hyper_param} compares different training hyper-parameters of \textbf{ST-$\textsc{SA}^2$} in the attraction domain of \textsc{FewShotWOZ} dataset.
We observe that the learning rate plays an essential role in training NLG models under the low-data setting.
If the learning rate is too large, the development loss may not converge because the training set is too small; if the learning rate is too small, the model may get stuck into the local optimal.
Finally, we find a good combination of learning rate and training epoch can help the model achieves the best performance, but the specific values vary across different domains.
We provide training hyper-parameter configurations of each domain in \autoref{sec:appendix-implementation}.
\section{Conclusions}
In this work, we present a two-phase self-augmentation self-training algorithm to deal with the few-shot dialogue generation problem in task-oriented dialogue systems.
We propose to select informative input MRs based on model's prediction uncertainty, and improve the pseudo response generation by aggregating randomly perturbed latent representations.
Empirical experiments on two few-shot NLG datasets show that our proposed method achieves the best performance among other baselines in both automatic and human evaluations. 
\section*{Limitations}


The performance of $\textsc{SA}^2$ self-training algorithm is influenced by the pre-trained language model used as the base generation model, because it offers the starting point for data selection and data augmentation.
Building a good pre-trained language model for the MR-to-Text generation task is non-trivial, but future work in this direction will certainly benefit few-shot learning on dialogue generation.
Besides, the $\textsc{SA}^2$ self-training algorithm requires large GPU resources for augmenting pseudo-labeled data.
A more computationally efficient decoding method of Transformer-based models would save a significant amount of time and GPU resources.

\section*{Acknowledgments}
The authors thank the anonymous reviewers for their useful comments and the UVa ILP group for helpful discussions. 
This work was supported by an Amazon Research Award to Yangfeng Ji.

\bibliography{anthology,custom}
\bibliographystyle{acl_natbib}

\clearpage
\appendix
\section{Details of $\textsc{SA}^2$ Self-training Algorithm}
\label{sec:appendix-framework}

We choose the pre-trained language model SC-GPT \citep{peng-etal-2020-shot} as our base generation model $p_\theta$.
We collect in-domain MRs from the training set of existing task-oriented dialogue datasets, such as MultiWOZ corpus \citep{budzianowski-etal-2018-multiwoz} and Schema-Guided Dialog corpus \citep{Rastogi_Zang_Sunkara_Gupta_Khaitan_2020}.
We use nucleus sampling \citep{Holtzman2020The} with the threshold $p=0.9$ to generate the output tokens for both synthetic text annotation and refined response generation.

\section{Dataset Details}
\label{sec:appendix-dataset}
Note that the original \textsc{FewShotWOZ} and \textsc{FewShotSGD} do not have a development set.
To create the standard training/dev/test data splits, we randomly sampled 10\% data from the original test set as the dev set, and kept the training set unchanged. 
For fair comparisons across different methods, we evaluated all methods on the newly split test set.
The detailed data statistics of \textsc{FewShotWOZ} is presented in \autoref{tab:data_woz}.
The detailed data statistics of \textsc{FewShotSGD} is demonstrated in \autoref{tab:data_sgd}.  
\begin{table*}[t]
  \centering
  \small
  \begin{tabular}{l|rrrrrrr}
    \toprule
     & \textbf{Restaurant} & \textbf{Laptop} & \textbf{Hotel} & \textbf{TV} & \textbf{Attraction} & \textbf{Train} & \textbf{Taxi} \\
    \midrule
    \# Training Pairs & 51 & 51 & 51 & 51 & 50 & 50 & 40 \\
    \# Dev Pairs & 12 & 137 & 7 & 68 & 34 & 65 & 4 \\
    \# Test Pairs & 117 & 1242 & 71 & 612 & 306 & 592 & 43 \\
    \# Unlabeled Data & 10,000 & 10,000 & 10,000 & 7,035 & 10,000 & 10,000 & 6,527 \\
    \bottomrule
  \end{tabular}
  \caption{\label{tab:data_woz}
  Data statistics for the original manual-labeled data $\mathcal{D}_L$ and the unlabeled data $\mathcal{D}_U$ on \textsc{FewShotWOZ}.
  }
\end{table*}

\begin{table*}[t]
  \centering
  \small
  \begin{tabular}{l|rrrrrrrr}
    \toprule
     & \textbf{Restaurants} & \textbf{Hotels} & \textbf{Flights} & \textbf{Buses} & \textbf{Events} & \textbf{Rentalcars} & \textbf{Services} & \textbf{Ridesharing} \\
    \midrule
    \# Training Pairs & 50 & 50 & 50 & 50 & 50 & 50 & 50 & 48 \\
    \# Dev Pairs & 961 & 401 & 272 & 427 & 836 & 287 & 793 & 819 \\
    \# Test Pairs & 8,657 & 3,615 & 2,453 & 3,845 & 7,526 & 2,592 & 7,146 & 7,378 \\
    \# Unlabeled Data & 10,000 & 10,000 & 10,000 & 10,000 & 10,000 & 10,000 & 10,000 & 8,259 \\
    \bottomrule
    \toprule
    & \textbf{Movies} & \textbf{Calendar} & \textbf{Banks} & \textbf{Music} & \textbf{Homes} & \textbf{Media} & \textbf{Travel} & \textbf{Weather} \\
    \midrule
    \# Training Pairs & 30 & 25 & 23 & 21 & 21 & 14 & 14 & 11 \\
    \# Dev Pairs & 737 & 532 & 332 & 732 & 563 & 568 & 528 & 193 \\
    \# Test Pairs & 6,634 & 4,793 & 2,988 & 6,594 & 5,073 & 5,121 & 4,753 & 1,742 \\
    \# Unlabeled Data & 7,604 & 5,355 & 3,343 & 7,347 & 5,657 & 5,703 & 5,299 & 1,947 \\
    \bottomrule
  \end{tabular}
  \caption{\label{tab:data_sgd}
  Data statistics for the original manual-labeled data $\mathcal{D}_L$ and the unlabeled data $\mathcal{D}_U$ on \textsc{FewShotSGD}.
  }
\end{table*}

\section{Experimental Details}
\label{sec:appendix-implementation}
\begin{table*}[t]
  \centering
  \small
  \begin{tabular}{llccccc}
    \toprule
    & \textbf{Domain} & \textbf{Epoch} & \textbf{LR} & $\text{\textbf{BLEU}}_{dev}$ & $\text{\textbf{BLEU}}_{test}$ & $\text{\textbf{ERR}}_{test}$ \\
    \midrule
    1 & Restaurant & 10 & 8e-7 & 38.10 & 36.48 & 2.60 \\
    2 & Laptop & 10 & 5e-6 & 34.19 &35.42 & 2.04  \\
    3 & Hotel & 10 & 1e-6 & 33.46 & 42.63 & 1.78 \\
    4 & TV & 10 & 1e-6 & 37.10 & 36.39 & 1.63 \\
    5 & Attraction & 20 & 5e-7 & 23.43 & 25.63 & 1.40 \\
    6 & Train & 10 & 8e-7 & 23.65 & 25.34 & 1.62 \\
    7 & Taxi & 10 & 1e-6 & 6.08 & 20.95 & 0.00 \\
    \bottomrule
  \end{tabular}
  \caption{\label{tab:train_config_woz}
  Training hyper-parameter configurations of \textbf{ST-$\textsc{SA}^2$} in \textsc{FewShotWOZ}, where \textbf{Epoch} is the number of training epochs within a self-training iteration, and \textbf{LR} is the initial learning rate at the beginning of each training epoch. We set the maximum self-training iteration $S=5$, and select the model which has the highest $\text{\textbf{BLEU}}_{dev}$ across all self-training iterations.
  }
\end{table*}

\begin{table*}[t]
  \centering
  \small
  \begin{tabular}{llcccc}
    \toprule
    & \textbf{Domain} & \textbf{Epoch} & \textbf{LR} & $\text{\textbf{BLEU}}_{dev}$ & $\text{\textbf{BLEU}}_{test}$ \\
    \midrule
    1 & Restaurants & 10 & 1e-6 & 20.69 & 20.42 \\
    2 & Hotels & 10 & 1e-6 & 22.69 & 22.90 \\
    3 & Flights & 10 & 5e-6 & 25.82 & 27.12 \\
    4 & Buses & 10 & 1e-6 & 21.74 & 21.16 \\
    5 & Events & 10 & 5e-6 & 26.46 & 25.32 \\
    6 & Rentalcars & 10 & 1e-5 & 20.67 & 20.70 \\
    7 & Services & 10 & 1e-6 & 28.57 & 28.34 \\
    8 & Ridesharing & 10 & 1e-6 & 23.61 & 23.28 \\
    9 & Movies & 10 & 1e-6 & 29.37 & 28.95 \\
    10 & Calendar & 10 & 1e-6 & 25.97 & 25.24 \\
    11 & Banks & 10 & 1e-6 & 27.45 & 28.14 \\
    12 & Music & 10 & 1e-6 & 27.06 & 27.23 \\
    13 & Homes & 10 & 1e-6 & 24.45 & 25.03 \\
    14 & Media & 10 & 1e-5 & 28.40 & 28.76 \\
    15 & Travel & 10 & 1e-6 & 24.09 & 25.34 \\
    16 & Weather & 10 & 5e-7 & 27.43 & 29.27 \\
    \bottomrule
  \end{tabular}
  \caption{\label{tab:train_config_sgd}
  Training hyper-parameter configurations of \textbf{ST-$\textsc{SA}^2$} in \textsc{FewShotSGD}, where \textbf{Epoch} is the number of training epochs within a self-training iteration, and \textbf{LR} is the initial learning rate at the beginning of each training epoch. We set the maximum self-training iteration $S=5$, and select the model which has the highest $\text{\textbf{BLEU}}_{dev}$ across all self-training iterations.
  }
\end{table*}
\paragraph{General Setups:}
The model is trained on an NVIDIA GeForce GTX 1080 Ti GPU server with 12GB memory.
For the learning rate, we use the linear rate scheduler with no warm-ups.
The AdamW optimizer \citep{loshchilov2018decoupled} with default weight decay is used to update the parameters.
For generation, we use nucleus sampling with $p=0.9$ across all experiments. 

\paragraph{SC-GPT:}
The pre-trained language model SC-GPT is loaded and fine-tuned on the original few-shot training set $\mathcal{D}_L$.
The training epoch is set to 10, the batch size is set to 1, and the initial learning rate is set to 1e-5 across all domains in both \textsc{FewShotWOZ} and \textsc{FewShotSGD}.

\paragraph{AUG-NLG:}
There are two learning stages.
In the first stage, the pre-trained language model SC-GPT is loaded and trained on the retrieved augmented data released by \citet{xu-etal-2021-augnlg}, where the training epoch is set to 10, the batch size is set to 4, and the initial learning rate is set to 1e-5 across all domains in both datasets.
In the second stage, the model checkpoint from the first stage is loaded and fine-tuned on the original few-shot training set $\mathcal{D}_L$, where the training epoch is set to 10, the batch size is set to 4, and the initial learning rate is set to 1e-5 across all domains in both datasets.

\paragraph{ST-ALL, ST-NLL, ST-$\textsc{SA}^2$:}
For all self-training methods, we start with the model checkpoint from the \textbf{SC-GPT} baseline.
The maximum self-training iteration is set to $S=5$.
For evaluation, we save all model checkpoints at each self-training iteration, and report the best-performed model which has the highest $\text{\textbf{BLEU}}_{dev}$ score among all iterations (not necessarily the last iteration).

For \textbf{ST-ALL} and \textbf{ST-NLL}, in each self-training iteration, the training epoch is set to 10, the batch size is set to 4, and the initial learning rate is set to 1e-5 across all domains in both datasets.

For the model hyper-parameters in \textbf{ST-$\textsc{SA}^2$}, we set $M=10$ in \autoref{eq:mean} and \autoref{eq:var}, and set $R=10$ in \autoref{eq:h_bar}.
For \textbf{ST-$\textsc{SA}^2$}, the training batch size is set to 4, and we report the detailed training epoch and initial learning rate across different domains and datasets for reproducibility purpose in \autoref{tab:train_config_woz} and \autoref{tab:train_config_sgd}.

\section{Self-Augmented Data Examples}
\label{sec:appendix-augment}
\begin{table*}[ht]
  \centering
  \small
  \begin{tabular}{p{0.3\textwidth}|p{0.63\textwidth}}
    \toprule
     \multicolumn{2}{c}{\textcolor{red}{\textbf{low}} $\mathbb{E}[p_\theta]$   \textcolor{red}{\textbf{low}} $Var[p_\theta]$} \\ \hline
    Input MR $\mathcal{D}_{U}$ & \textit{inform ( choice = several ) @ request ( area = ? ) } \\
    Self-augmented data $\mathcal{D}_A$ (\textbf{Phase I}) & i have several restaurants that are good for lunch or dinner \\
    Pseudo-labeled data $\mathcal{D}_{L'}$ (\textbf{Phase II}) & there are several restaurants that meet your needs \\
    \midrule
    \multicolumn{2}{c}{\textcolor{red}{\textbf{low}} $\mathbb{E}[p_\theta]$   \textcolor{teal}{\textbf{high}} $Var[p_\theta]$} \\ \hline
    Input MR $\mathcal{D}_{U}$ & \textit{inform ( choice = several ) @ request ( area = ? ) } \\
    Self-augmented data $\mathcal{D}_A$ (\textbf{Phase I}) & there are several restaurants that match your criteria \\
    Pseudo-labeled data $\mathcal{D}_{L'}$ (\textbf{Phase II}) & we have several restaurants that fit your criteria \\
    \midrule
    \multicolumn{2}{c}{\textcolor{teal}{\textbf{high}} $\mathbb{E}[p_\theta]$   \textcolor{red}{\textbf{low}} $Var[p_\theta]$} \\ \hline
    Input MR $\mathcal{D}_{U}$ & \textit{request ( area = ? ) } \\
    Self-augmented data $\mathcal{D}_A$ (\textbf{Phase I}) & what is the area you looking for \\
    Pseudo-labeled data $\mathcal{D}_{L'}$ (\textbf{Phase II}) & what is the area you looking for \\
    \midrule
    \multicolumn{2}{c}{\textcolor{teal}{\textbf{high}} $\mathbb{E}[p_\theta]$   \textcolor{teal}{\textbf{high}} $Var[p_\theta]$} \\ \hline
    Input MR $\mathcal{D}_{U}$ & \textit{inform ( choice = several ) @ request ( area = ? ) } \\
    Self-augmented data $\mathcal{D}_A$ (\textbf{Phase I}) & there are several restaurants in the area you'd like to dine in \\
    Pseudo-labeled data $\mathcal{D}_{L'}$ (\textbf{Phase II}) & i have several restaurants that meet your criteria. what is the area you looking for \\
    \bottomrule
  \end{tabular}
  \caption{\label{tab:self_aug_examples}
  Examples of self-augmented data $\mathcal{D}_A$ and pseudo-labeled data $\mathcal{D}_{L'}$ under different data selection strategies in the \textbf{Restaurant} domain of \textsc{FewShotWOZ}.
  }
\end{table*}

\autoref{tab:self_aug_examples} shows some examples of self-augmented data $\mathcal{D}_A$ and pseudo-labeled data $\mathcal{D}_{L'}$ under different data selection strategies in the \textbf{Restaurant} domain of \textsc{FewShotWOZ}.

\section{Model Prediction Examples}
\label{sec:appendix-generation}
\begin{table*}[ht]
  \centering
  \small
  \begin{tabular}{p{0.12\textwidth}|p{0.8\textwidth}}
    \toprule
    \multicolumn{2}{c}{\textbf{Domain: Ridesharing}} \\ \hline
    Input MR & \textit{INFORM ( ride\_fare = \$11.02 ; approximate\_ride\_duration = 20 )} \\
    Human & it should take around 20 minutes and cost \$11.02. \\
    \textbf{SC-GPT} & the cost of the cab is \$11.02 and the ride lasts 20 minutes. \\
    \textbf{AUG-NLG} & the ride lasts 20 minutes and will cost \$11.02. \\
    \textbf{ST-ALL} & it will take 20 minutes and cost \$11.02. \\
    \textbf{ST-NLL} & the cost of the cab is \$11.02 and the ride will last 20 minutes. \\
    \textbf{ST-$\textsc{SA}^2$} (ours) & the cost is \$11.02 and it will take 20 minutes. \\
    \midrule
    \multicolumn{2}{c}{\textbf{Domain: Media}} \\ \hline
    Input MR & \textit{OFFER ( title = the visitor, viva las vegas )} \\
    Human & how about the visitor or viva las vega? \\
    \textbf{SC-GPT} & the visitor, viva las vegas is the only movie that meets your criteria. \\
    \textbf{AUG-NLG} & the visitor viva las vegas. \\
    \textbf{ST-ALL} & the visitor, viva las vegas is also available. \\
    \textbf{ST-NLL} & the visitor, viva las vegas is playing. \\
    \textbf{ST-$\textsc{SA}^2$} (ours) & how about the visitor, viva las vegas? \\
    \midrule
    \multicolumn{2}{c}{\textbf{Domain: Hotels}} \\ \hline
    Input MR & \textit{CONFIRM ( where\_to = sydney ; number\_of\_adults = 1 ; check\_in\_date = next monday ; check\_out\_date = march 11th )} \\
    Human & please confirm: the location is sydney for a group of 1. the check in date will be next monday and the check out date will be march 11th. \\
    \textbf{SC-GPT} & please confirm the following: you'd like to stay in sydney for 1 person from next monday to march 11th. \\
    \textbf{AUG-NLG} & this is for a hotel in sydney. the reservation is for 1 person. the check-in date is next monday. \\
    \textbf{ST-ALL} & sure. how about sydney? there's 1 person in your group, and you'll be arriving next monday. you'll be leaving march 11th. \\
    \textbf{ST-NLL} & okay, i've got that. you 'd like a 1 person reservation for a house in sydney that 'd be held next monday, and will be'retired on march 11th. is that right? \\
    \textbf{ST-$\textsc{SA}^2$} (ours) & so to recap: you want a house in sydney for 1 person with a departure date of next monday and a march 11th check out. \\
    \bottomrule
  \end{tabular}
  \caption{\label{tab:model_outputs_sgd}
  Examples of model generation results from the test set of \textsc{FewShotSGD}.
  }
\end{table*}

\begin{table*}[ht]
  \centering
  \small
  \begin{tabular}{p{0.12\textwidth}|p{0.8\textwidth}}
    \toprule
    \multicolumn{2}{c}{\textbf{Domain: Restaurant}} \\ \hline
    Input MR & \textit{inform\_no\_match ( pricerange = expensive ; area = bernal heights )} \\
    Human & there are no expensive restaurants in bernal heights \\
    \textbf{SC-GPT} & there are no expensive restaurants in the bernal heights area \\
    \textbf{AUG-NLG} & unfortunately there is no restaurant near bernal heights that is expensive \\
    \textbf{ST-ALL} & there is no restaurant in the expensive price range in the bernal heights  \\
    \textbf{ST-NLL} & unfortunately there is no restaurants in the expensive price range in the bernal heights area \\
    \textbf{ST-$\textsc{SA}^2$} (ours) & there are no expensive restaurants in the bernal heights area \\
    \midrule
    \multicolumn{2}{c}{\textbf{Domain: Laptop}} \\ \hline
    Input MR & \textit{inform\_only\_match ( name = tecra morpheus 65 ; type = laptop ; batteryrating = exceptional ; pricerange = moderate )} \\
    Human & the tecra morpheus 65 laptop is a moderate price range computer, with an exceptional battery rating \\
    \textbf{SC-GPT} & tecra morpheus 65 laptop is an exceptional laptop in the moderate price range \\
    \textbf{AUG-NLG} & tecra morpheus 65 is a nice laptop that is in the moderate price range. it has exceptional battery rating \\
    \textbf{ST-ALL} & tecra morpheus 65 is a laptop in the moderate price range and comes with an exceptional battery rating \\
    \textbf{ST-NLL} & the tecra morpheus 65 is a laptop in the moderate price range with exceptional battery rating \\
    \textbf{ST-$\textsc{SA}^2$} (ours) & tecra morpheus 65 is a laptop in the moderate price range with an exceptional battery rating \\
    \midrule
   \multicolumn{2}{c}{\textbf{Domain: Hotel}} \\ \hline
    Input MR & \textit{inform\_count ( type = hotel ; count = 182 ; hasinternet = dont\_care )} \\
    Human & there are 182 hotels if it does not matter if they have internet or not \\
    \textbf{SC-GPT} & there are 182 hotels that do not care if there internet is \\
    \textbf{AUG-NLG} & there are 182 hotels that do not care about internet access \\
    \textbf{ST-ALL} & there are 182 hotels that do not care if they have internet or wi-fi \\
    \textbf{ST-NLL} & there are 182 hotels that do not care if they have internet or not \\
    \textbf{ST-$\textsc{SA}^2$} (ours) & there are 182 hotels that do not care about internet \\
    \midrule
    \multicolumn{2}{c}{\textbf{Domain: TV}} \\ \hline
    Input MR & \textit{inform\_no\_match ( type = television ; hasusbport = true ; screensizerange = medium ; family = l6 )} \\
    Human & i apologize but there are no televisions with medium screen size , that has usb ports \\
    \textbf{SC-GPT} & no, there is no television in the l6 family with usb ports that is a medium screen size \\
    \textbf{AUG-NLG} & the l6 television has usb ports, whereas the television without has only a medium screen size. \\
    \textbf{ST-ALL} & there are no televisions that are in the medium family, with usb ports or in the l6 product family. \\
    \textbf{ST-NLL} & there are no televisions with usb ports in the medium size family or l6 family. \\
    \textbf{ST-$\textsc{SA}^2$} (ours) & there are no televisions in the medium screen size range with usb ports in the l6 family. \\
    \midrule
    \multicolumn{2}{c}{\textbf{Domain: Attraction}} \\ \hline
    Input MR & \textit{sorry ( area = centre ; choice = none )} \\
    Human & i have none available in the centre. would you like a different area or attraction? \\
    \textbf{SC-GPT} & i don't have any results for centre. would you like me to look in any other areas? \\
    \textbf{AUG-NLG} & there aren't any listings for centre. would you like to try a specific area? \\
    \textbf{ST-ALL} & there are none in the centre. would you like me to check other areas? \\
    \textbf{ST-NLL} & there are none in the centre. would you like me to check in another area? \\
    \textbf{ST-$\textsc{SA}^2$} (ours) & sorry, i don't have any listings for centre. do you want me to check another area? \\
    \midrule
    \multicolumn{2}{c}{\textbf{Domain: Train}} \\ \hline
    Input MR & \textit{inform ( leaveat = 16:34 ; id = tr8132 ) @ request ( people = ? )} \\
    \textit{Human} & i have the tr8132 which leaves at 16:34. how many tickets please? \\
    \textbf{SC-GPT} & tr8132 leaves at 16:34. how many tickets will you need? \\
    \textbf{AUG-NLG} & there is a train that leaves at 16:34. how many tickets would you like? \\
    \textbf{ST-ALL} & i have train tr8132 leaving at 16:34. how many tickets would you like? \\
    \textbf{ST-NLL} & i have train tr8132 that leaves at 16:34. how many tickets would you like to book? \\
    \textbf{ST-$\textsc{SA}^2$} (ours) & tr8132 leaves at 16:34. how many tickets would you like? \\
    \midrule
    \multicolumn{2}{c}{\textbf{Domain: Taxi}} \\ \hline
    Input MR & \textit{request ( destination = ? ; arriveby = ? )} \\
    Human & i would be happy to book your taxi but i will need your destination and preferred arrival time as well. \\
    \textbf{SC-GPT} & sure, where would you like to travel to? also where would you like to be picked up from? \\
    \textbf{AUG-NLG} & what time would you like to arrive? \\
    \textbf{ST-ALL} & sure, where would you like to go to and what time? \\
    \textbf{ST-NLL} & what is your destination? do you have a arrival time in mind? \\
    \textbf{ST-$\textsc{SA}^2$} (ours) & sure, where would you like to go to? also, what time would you like to arrive? \\
    \bottomrule
  \end{tabular}
  \caption{\label{tab:model_outputs_woz}
  Examples of model generation results from the test set of \textsc{FewShotWOZ}.
  }
\end{table*}

\autoref{tab:model_outputs_sgd} demonstrates some examples of model generation results in \textsc{FewShotSGD}.
\autoref{tab:model_outputs_woz} demonstrates some examples of model generation results in \textsc{FewShotWOZ}.

\end{document}